# Using Anatomical Markers for Left Ventricular Segmentation of Long Axis Ultrasound Images

Yael Petrank*, Nahum Smirin*, Yossi Tsadok*, Zvi Friedman[†], Peter Lysiansky[†], Dan Adam*

*Abstract*— Left ventricular segmentation is essential for measuring left ventricular function indices. Segmentation of one or several images requires an initial guess of the contour. It is hypothesized here that creating an initial guess by first detecting anatomical markers, would lead to correct detection of the endocardium. The first step of the algorithm presented here includes automatic detection of the mitral valve. Next, the apex is detected in the same frame. The valve is then tracked throughout the cardiac cycle. Contours passing from the apex to each valve corner are then found using a dynamic programming algorithm. The resulting contour is used as an input to an active contour algorithm. The algorithm was tested on 21 long axis ultrasound clips and showed good agreement with manually traced contours. Thus, this study demonstrates that detection of anatomic markers leads to a reliable initial guess of the left ventricle border.

## I. INTRODUCTION

Echocardiographers are required to assess myocardial function, thus in nearly all echo studies they need to measure indices of global left ventricular function, such as left ventricular volumes, ejection fraction, and endocardial shortening. In order to measure these indices, segmentation of the left ventricular (LV) wall in 2 dimensional (2D) ultrasound images must be performed, either by eyeballing or by semi-automatic or automatic techniques. The latter group of techniques are mandatory when objectivity is important, and

Dr. Yael Petrank is with the Department of Biomedical Engineering, Technion-IIT, Technion City, Haifa 32000, Israel. (E-mail: yaelp@tx.technion.ac.il).
Dr. Nahum Smirin is with the Department of Biomedical Engineering, Technion-IIT, Technion City, Haifa 32000, Israel. (E-mail: nsmirin@gmail.com ).
Dr. Yossi Tsadok is with the Department of Biomedical Engineering, Technion-IIT, Technion City, Haifa 32000, Israel. (E-mail: yossitsadok@gmail.com ).
Dr. Zvi Friedman is with GE Healthcare Inc., Haifa, Israel. (E-mail: zvi.friedman@ge.com ).
Dr. Peter Lysyansky is with GE Healthcare Inc., Haifa, Israel. (E-mail: peter.lysyansky@ge.com ).
Prof. Dan Adam is with the Department of Biomedical Engineering, Technion-IIT, Technion City, Haifa 32000, Israel. (E-mail: dan@bm.technion.ac.il ).

are essential for real time calculation of these indices. Additionally, the curve describing the myocardial endocardial border can also be used for registration between myocardial data sets obtained from different modalities. It is also essential for objective operator-independent segmentation of the LV wall for measurements of local myocardial function, e.g. speckle tracking based 2D strain or layer specific strain analysis.

Previous reports [1-9, 11-14] have presented several automatic or semi-automatic algorithms for left ventricular segmentation that were designed for long axis ultrasound images. Among them is an early study by Friedland and Adam [1] who used an optimization method, simulated annealing, to detect the endocardial border. As an initial guess an ellipse was used, where its parameters were estimated from locations of gradients found along rays originating from the center of mass of the left ventricle. Herlin et al [2,3] developed a spatial–temporal model for segmenting cardiac ultrasound sequences. The parameters of the energy function were determined from a manually segmented frame. Mikic et al [4] developed an active contour algorithm for detecting the endocardial border as well as the mitral valve leaflet. The algorithm requires an initial manual segmentation of the borders on one frame. Sun et al [5] developed a segmentation scheme in which the heart dynamics is studied from a training set of data. Coppini et al [6] used a Laplacian-of –Gaussian filter to detect edges in ultrasonic images, then used neural networks to extract the endocardial border and eventually to reconstruct the endocardial surface in 3-D. Mignotte et al [7] used deformable templates for segmentation of the endocardium. Orderud et al [8,9] developed a framework for real time left ventricular tracking of 3D ultrasonic data. Initially, a truncated ellipsoid model was used with deformation parameters for translation, orientation, scaling and bending in all dimensions. The tracking framework automatically detected initial left ventricular position and tracked the motion and shape changes throughout the cycle.

In recent years, the level set frame work, presented initially by Osher and Sethian [10] has been used in algorithms for left ventricular segmentation. Lin et al [11] used a multi-scale level set framework for segmentation of long axis ultrasonic images in which the valve was closed. Corsi et al [12] used the level set technique to segment 3D echocardiographic data. The initial guess is a manual segmentation by the physician.

Paragios [13] proposed a segmentation algorithm that uses the level set approach. The forces acting on the contour depend both on visual information and on prior shape knowledge. Angelini et al [14] used the level set frame work for left ventricular segmentation in three dimensional data. The shape of the left ventricle was initialized as a cone with ellipsoidal cross section.

One of the main disadvantages of using active contour algorithms is that these algorithms use an iterative process until they converge, and that they require an initial guess of the contour. If the initial contour is created by manual tracing of the endocardium, then this process is too time-consuming. Additionally, since in many cases the initial guess is not close to the endocardial border, e.g. when an elliptical model of the left ventricle is used, the curve will probably not converge in real time, or will be strongly influenced by the initial model. Specifically in cardiac images, there is a chance that the curve will attach to anatomical features (such as the mitral valve) which can misdirect the desired curve, in case the initial guess is far from the endocardial border.

The aim of the present work was to develop an algorithm for creating an initial guess for the endocardial border, which would be as close as possible to the desired contour. In order for the initial guess to be robust and accurate, much effort has been put into developing sub algorithms for the detection of the valve corners and the apex of the left ventricle, as it was assumed that correct detection of those anatomical markers would lead to correct detection of the endocardial border. After creating the initial guess, it was used as an input to an active contour algorithm in a level set frame work. The active contour algorithm serves only as a fine tuning step, with a limited number of iterations, to allow a better attachment between the curve and the myocardium. The active contour algorithm used for this step is based on a study by Lankton and Tannenbaums [15], who developed an active contour algorithm in which the energy depends on the gray level statistics and is calculated locally for each point on the curve. Each point is then advanced in the direction of minimizing this energy. This method is suitable for segmentation of cardiac ultrasound data because of the large heterogeneity of image statistics around the endocardial border.

The steps of our algorithm are described in the second section. Results are presented in the third section and discussed in the fourth section.

## II. MATERIALS AND METHODS

Data for our study was obtained at, and provided by the Dept. of Cardiology, Pneumology and Angiology, at the University Hospital RWTH Aachen, Germany, and by the Dept. of Cardiology at the Oslo University Hospital, Rikshospitalet, Oslo, Norway with Informed consent for use of images approved by the ethics committee.

The algorithm presented here may function completely automatically. The user is allowed, though, to manually correct the positions of the valve corners and the apex, thus a maximum of three manually marked points may be needed for a whole ultrasonic sequence. Figure 1 exhibits the flow chart of the segmentation scheme. A detailed description of each step appears below.

**Step 1: Automatic detection of Mitral valve corners in one frame**:

The first step of the algorithm comprises of locating automatically the septal and lateral corners of the Mitral valve, (or in other words, finding the location of the left ventricle base), in one frame at systole. During the systolic phase the mitral valve is closed, so that pixels that comprise the bridge-shaped valve may easily be detected. These are characterized as pixels with high gray level values, with a column of low gray level pixels above them (when considering an apical view, in which the valve is oriented horizontally). In Figure 2, the pixels that mark the valve are colored in green. The detected valve pixels may cover only part of the valve, but knowing that those pixels are located on the valve allows to define a new location above one of those pixels, but inside the ventricular blood pool. From this location a search is made, first towards the lateral wall and then towards the septal wall (Figure 3), where pixels with a significant higher gray level are sought. When such an increase in gray level is encountered, it is assumed that the lateral or septal walls are located. Regions of interest in which the valve corners are located are now defined (figure 4). Each corner is then detected in its corresponding ROI, by locating pixels that their surroundings resemble a bright "L" shape (for the septal corner, and its mirror image for the lateral corner) on a dark background. This step is done by first defining binary "L" shaped masks. Then, for each pixel in each region of interest, a sub image, surrounding that pixel and in the size of the mask, is selected. The sub image is transformed into a binary mask by giving the value of 1 to pixels with gray level exceeding a threshold, and the value of 0 to pixels with a lower gray level. The similarity of the sub-image to the binary mask is tested by calculating the percent of non-identical pixels. If this percentage is lower than a given threshold, the pixel is considered a "corner pixel". This process results in several corner-pixels per corner. The last procedure appears as a diagram in figure 5. The final corner location is the mean location of these pixels. Figure 6 exhibits an example of the corner pixels (a) and the final corner locations (b).

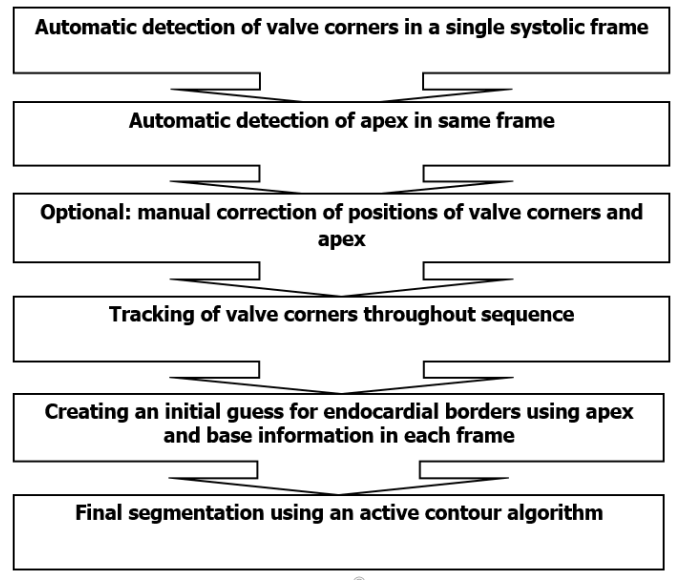

**Figure 1:** flow chart of segmentation algorithm

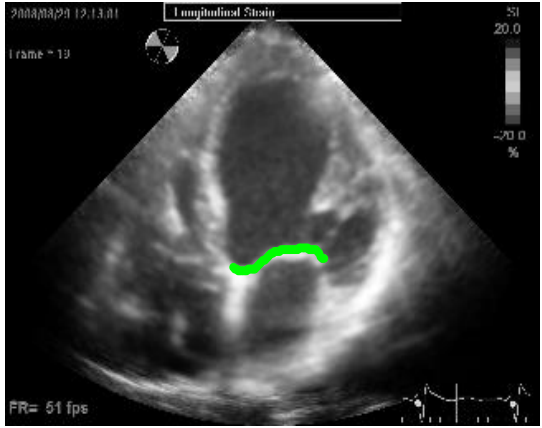

**Figure 2**: Valve pixels

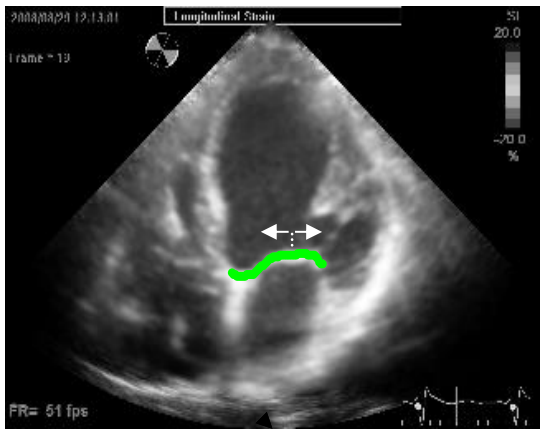

**Figure 3**: location above one valve pixel from which a point is advanced towards the right until it hits the lateral wall, and then to the left, until it hits the septal wall.

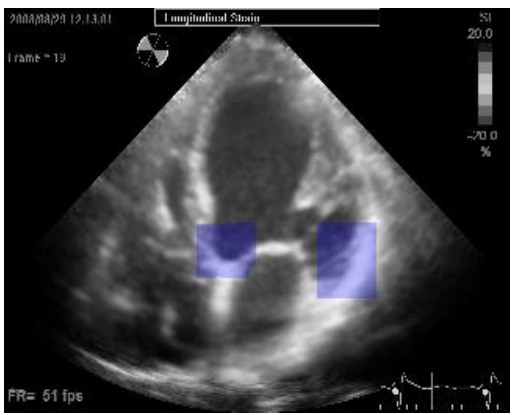

**Figure 4**: region of interests for left and right valve corners

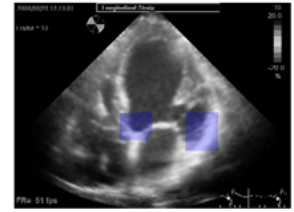

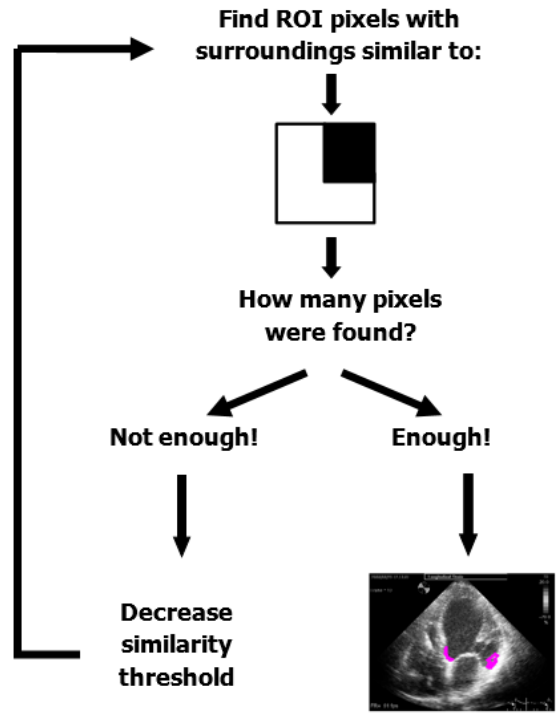

**Figure 5:** A diagram describing the search for valve corner pixels (for the left corner an "L" shaped mask is used, and its mirror image is used for the right corner)

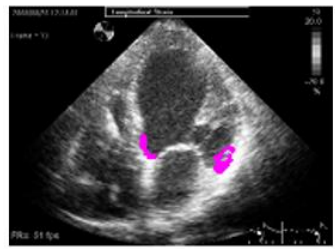 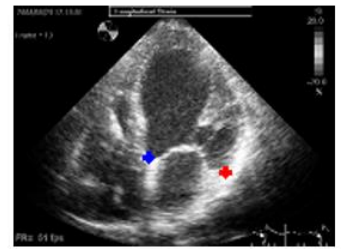

(a)  (b)

**Figure 6:** Pixels whose surroundings resemble a corner mask (a), and the mean location of those pixels, which defines the corner location (b)

**Step 2: Apex detection**

When acquiring long axis views in transthoracic echocardiography, the ultrasound beam passes through relatively static body layers until it reaches the apex of the heart. Consequently, correlating the gray level pattern of a small section of the image located in the static region with a matching section in the following image will yield a high correlation value. When performing the same operation but in layers of the myocardium, there is a drop in the correlation values as a result of the heart motion. A threshold value of the estimated correlation can be set, which indicates the epicardial border.

The detection of the apex is performed for a single systolic frame. It is initiated by defining several direction lines and sub direction lines which originate from the base of the left ventricle. The base location and orientation have been already determined in step 1. Figure 7 exhibits the main direction lines (green), originating from the midpoint between the valve corners, and the sub direction lines (magenta) defined for one of the main lines.

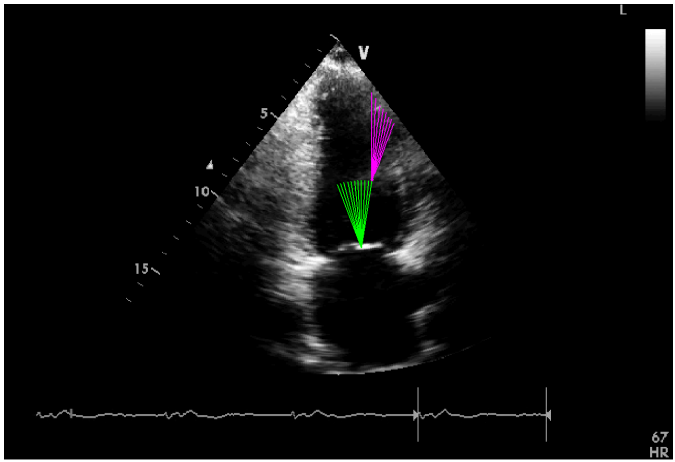

**Figure 7:** main direction lines (green) and sub direction lines defined for one of the main direction lines (magenta).

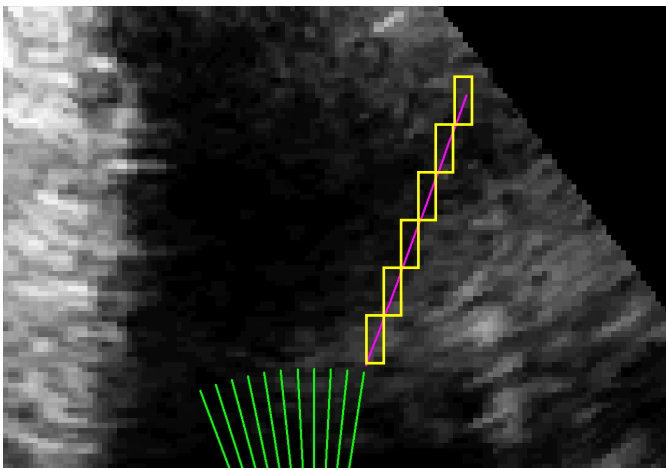

**Figure 8**: image blocks placed along a sub line

In the next step, blocks are defined around pixels positioned along each sub line. An example of blocks placed along a sub line is shown in figure 8. Each block is correlated with a matching block in the following frame. The "neighbor" block is correlated with its other neighbor, and so on. The mean of all the correlation values is calculated for each block. The epicardium is defined as the location in which the mean correlation value drops from a high value, corresponding to external layers, to below a threshold value (sub lines which do not have this pattern are excluded). During the search, several epicardial points are found, for each of the rays. The apical epicardial point is the one that is farthest from the base midpoint, as shown in figure 9.

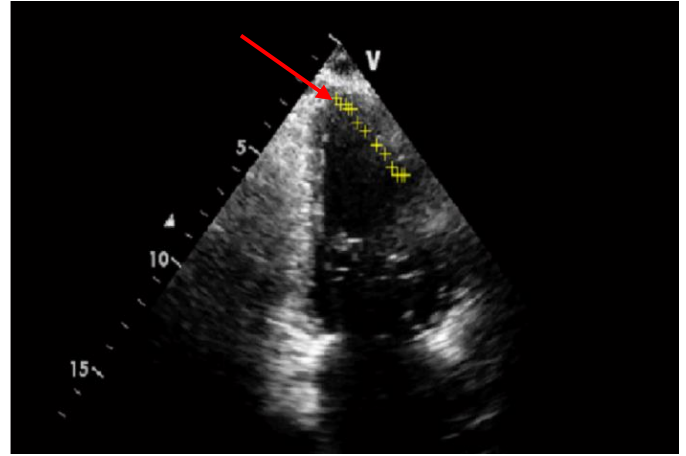

**Figure 9:** epicardial apex point is the farthest epicardial point from the base

If there exists a significant gray level gradient along the line connecting the apex and mid base, and close to the apex, it is assumed to be the border between the apical endocardium and the blood. Otherwise, the estimated location of the endocardial apex is a few millimeters closer to the base on that line.

**Step 3: Tracking of mitral valve corners in all frames**

The next step is tracking the location of the valve corners throughout all the frames in the sequence: as shown in figure 10, two square reference image blocks are defined for each valve corner, one surrounding the corner and a smaller one, slightly shifted and elevated from it. In the next frame, a search for similar blocks is carried out within a search region, separately for the septal and the lateral corners. For each pixel in the search region, two square blocks, positioned in the same manner as the reference squares, are defined. The sum of absolute difference (SAD) is calculated for larger square and then for the other, as described in equations 1 and 2:

$$SAD_1 = \sum_{i=1}^{n1}\sum_{j=1}^{m1}\left|I1_{ref}(i,j) - I1_{next}(i+k,j+l)\right| \quad (1)$$

$$SAD_2 = \sum_{i=1}^{n2}\sum_{j=1}^{m2}\left|I2_{ref}(i,j) - I2_{next}(i+k,j+l)\right| \quad (2)$$

Where I1ref and I2ref are the two reference image blocks within one image, sized (n1xm1) and (n2xm2) pixels respectively, I1next and I2next are matching blocks within the next image, k,l are the x and y shifts in the block positions within the next image, and are included in the search region.

The combined SAD result, for each pixel in the search region, is a multiplicity of the two SAD values, where one SAD result is squared for weight amplification:

$$SAD_{12} = SAD_1^2 \times SAD_2 \quad (3)$$

For each valve corner, the location in the next frame, which yields the best result, is considered the new corner location. The reference regions are then replaced by the best matching squares and the tracking continues to the next frame.

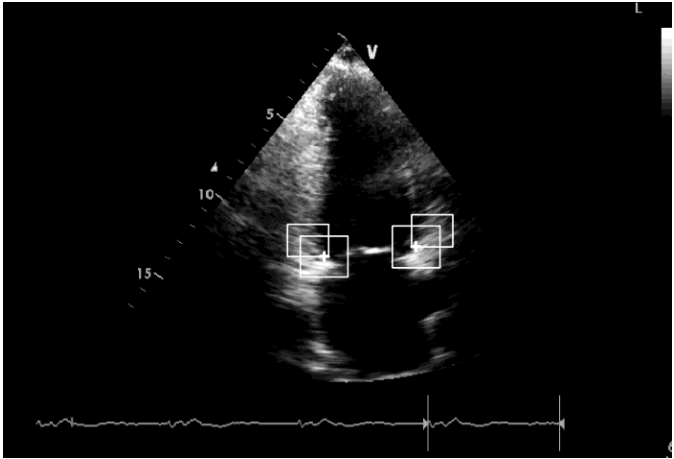

**Figure 10:** reference regions for the right valve corner (red, magenta) and the left valve corner (blue, green)

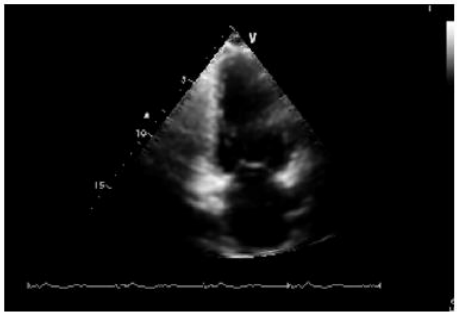

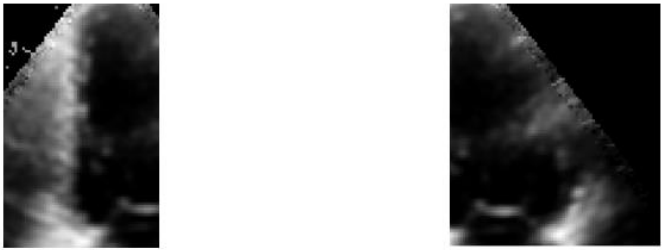

**Figure 11:** in order to create an initial guess for the left ventricular border, the left ventricle (top) is filtered and then partitioned into left and right parts (bottom).

**Step 4: Initial guess of left ventricular borders in all frames**

The current step uses dynamic programming (DP) for creating an initial guess of the endocardial border. DP is a method in which the optimal global solution of a problem can be found, and can be used in many different applications [16-24]. Before the DP algorithm is applied, each image is partitioned into the left and right parts of the left ventricle. The left part includes the apex and septal corner of the valve, and the right part includes the apex and lateral corner of the valve, as shown in figure 11. The two images undergo processing so that noise and texture features that may cause errors in the detection of the endocardial border - are eliminated. The gradient image of each part is then created. The contour of the endocardium is found for each part of the image using the DP algorithm. In general, the algorithm uses a cost function that depends on gray level gradients, continuity and smoothness of the curve, to find the most likely path in the gradient image space. This path is forced to start at the apex point and end at the base (septal or lateral corners of the valve, depending which part of the image is being processed). A detailed description of the algorithm appears in the appendix. The gradient image and the resulting path, superimposed on the image, are shown in figure 12, a and b. The left and right paths are then joined and smoothed to form the endocardial border contour. Figure 13 exhibits the joined left and right paths, before (a) and after (b) smoothing.

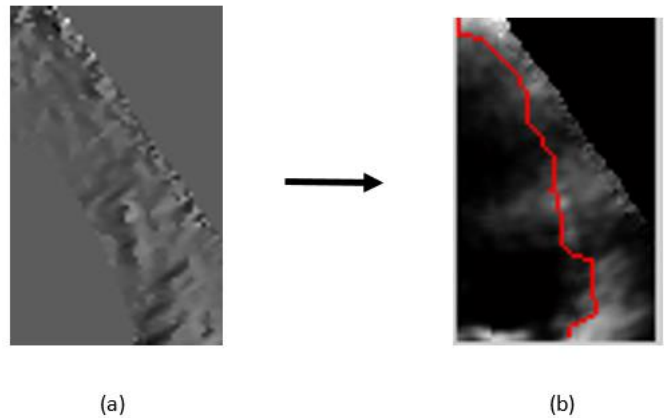

**Figure 12:** gradient image (a) and the resulting path, superimposed on the image (b)

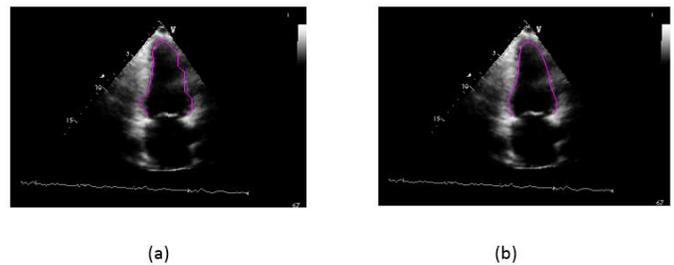

**Figure 13:** Joint left and right paths before (a) and after (b) smoothing

**Step 5: Final segmentation of all frames using an active contour algorithm**

In the fifth and final step of the algorithm, the results of step 4 are used as an initial guess for an active contour algorithm. The aim of this step is to allow a better attachment between the contour and the endocardial border. The algorithm is based on Lankton and Tannenbaum's work [15]. It uses the local gray level statistics around each contour point to evolve the curve and achieve better separation between gray levels inside and outside the contour at that point, and as a result, better attachment of the curve to the endocardial border. This method is well suited to ultrasound images, as gray level statistics vary greatly along the endocardial border.

### III. RESULTS

A set of 21 different ultrasound clips, with a total of 749 frames was segmented. Results of automatic detection of the mitral valve, from several different clips, are exhibited in Fig. 14 and the results of automatic detection of the apex are exhibited in Fig. 15. The results of the contour that serves as an initial guess, in one clip, are exhibited for every other frame in Fig. 16, and the results of the final segmentation are exhibited for the same frames in Fig. 17.

In order to measure the ability of the algorithm to correctly follow the endocardial borders, all frames were manually traced by an experienced viewer. The percent error of the final contour found automatically was defined as the non-overlapping areas between the automated and manual contours, as compared with the area enclosed by the manual contour. Errors were calculated for the contours that served as the initial guess and those that served for the final segmentation, after the stage of the fine tuning, which was performed by the active contour method.

The mean error for all frames was 13.2 ±4.0% for the contours that served as the initial guess, and 12.8±4.1% for the contours after the fine tuning stage. The mean error for the contours found automatically (initial guess stage) presented in Fig. 16 was 12.9±2.3% and was 12.7±2.2% for the contours presented in Fig. 17 (fine tuning stage). The results indicated that in most frames the contour found automatically successfully followed the endocardial border.

### IV. DISCUSSION

In this study, a novel method is presented for the segmentation of ultrasound images of the long axis cross section. There are numerous papers describing the segmentation of the LV myocardial tissue, since there is a significant clinical need for such a tool. Unfortunately, the results of these methods have not been accurate enough to allow routine clinical usage. The hypothesis of the current study was that the results of segmentation of the cardiac LV tissue depend on the initial guess or initial geometric model being used, which in most cases are manually drawn. Manual tracing is not objective, and different traces drawn by different users may lead to different classification results. Manual tracing is also time-consuming. The usage of a geometric model as the initial guess may require numerous iterations before achieving convergence to the true shape of the left ventricle. In the latter case, the contour may also attach to anatomical features inside the ventricle that are not part of the LV walls (e.g. the leaflets of the open valve, papillary muscles). The emphasis of the present study was on creating a reliable initial contour of the left ventricle that will then be used as an initial guess for an active contour algorithm. It was assumed that a correct detection of the endocardial border would depend on a correct detection of anatomical markers on the borders: the mitral valve and apex.

The need for an objective automatic initial guess of the left ventricle cavity is obvious for time saving, and identifying the locations of the valve corners and apex specifically, is a big advantage. This is confirmed in the segmentation result displayed in Friedland and Adam's work [1], where simulated annealing was used for segmentation of the left ventricle, but without identifying those locations. It is noticeable that in some frames, the contour of the left ventricle crosses into the left atrium. In our algorithm, in addition to the left ventricular contour, the valve corners locations in all frames are also inputs to the active contour algorithm. They are used to limit the evolution of the curve at the valve area, so that it will not cross into the left atrium even when the valve is open. The need for the information regarding the mitral valve location is also demonstrated in the works by Herlin et al [2,3]. They developed a model that depends on both the spatial properties and the temporal properties for segmenting cardiac ultrasound sequences. As mentioned in [2], initially, their model did not depend on temporal information and as a result the cardiac cavity was not defined accurately when mitral valve was open. The problem was solved by adding a term to their model, according to which the probability that a pixel will be considered an inside-the-cavity pixel will be higher if it was considered so in the previous and next frame. This constraint solved the problem of leakage into the atrium or ventricle (depending on which cavity was being segmented) when the valve was open, but the drawback was that the segmentation result became very stable and some information about the cavity shape changes during the cycle was lost [3].

In the long axis view, the apex region suffers from artifacts, such as clutter. There are studies, such as [25] which attempt to suppress those phenomena, but there is always the danger of suppressing the apical myocardium as well. Our apex detection algorithm depends on time domain behavior and not on image properties, and therefore, like the valve corners, can be used as a constraint for the segmentation curve, in case it will be erroneous due to the clutter.

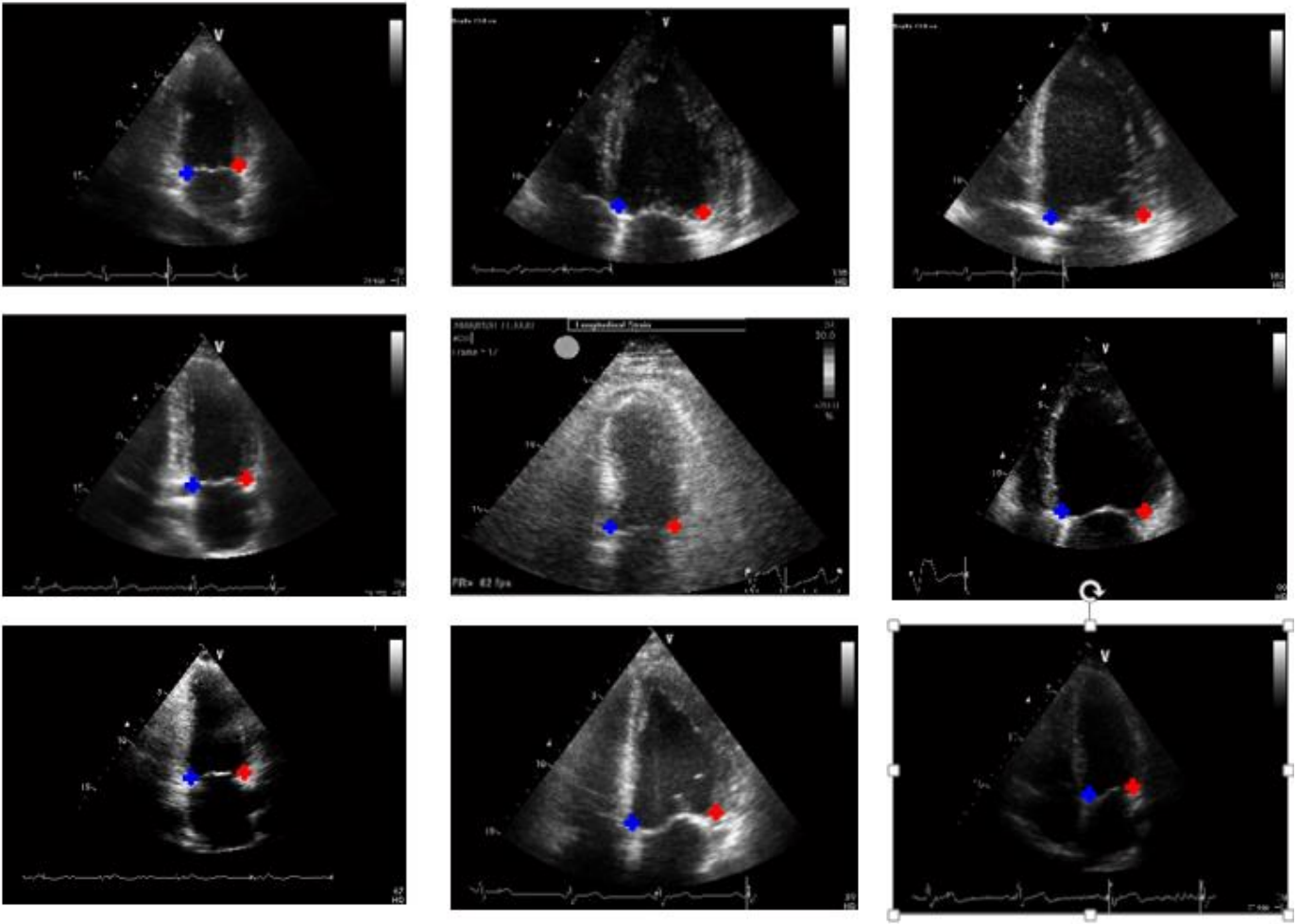

**Figure 14:** Mitral valve corners found automatically

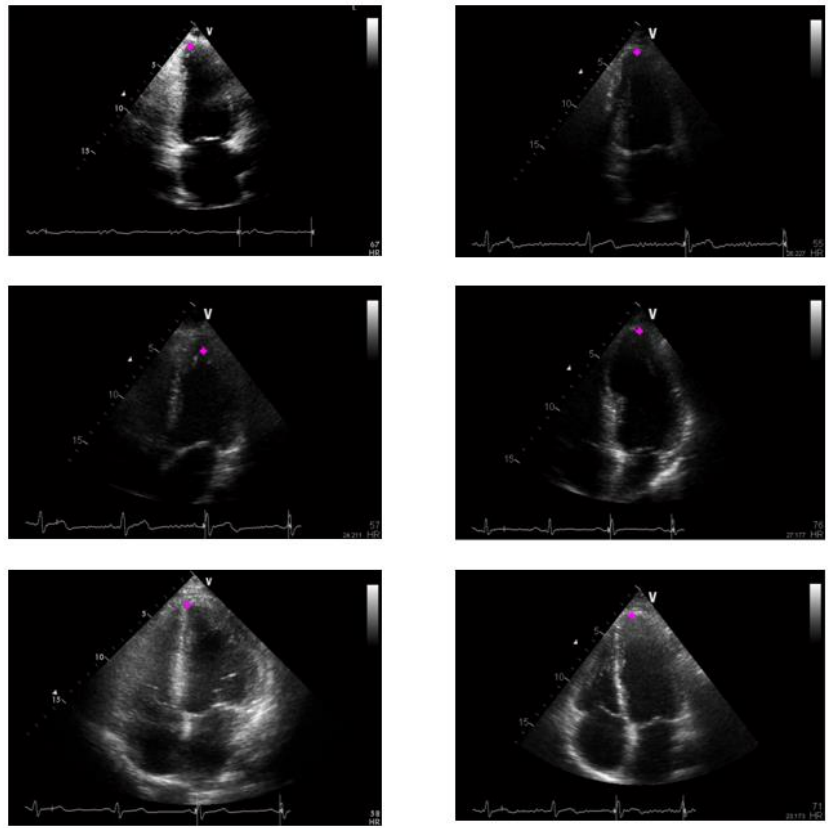

**Figure 15:** The apical location found automatically

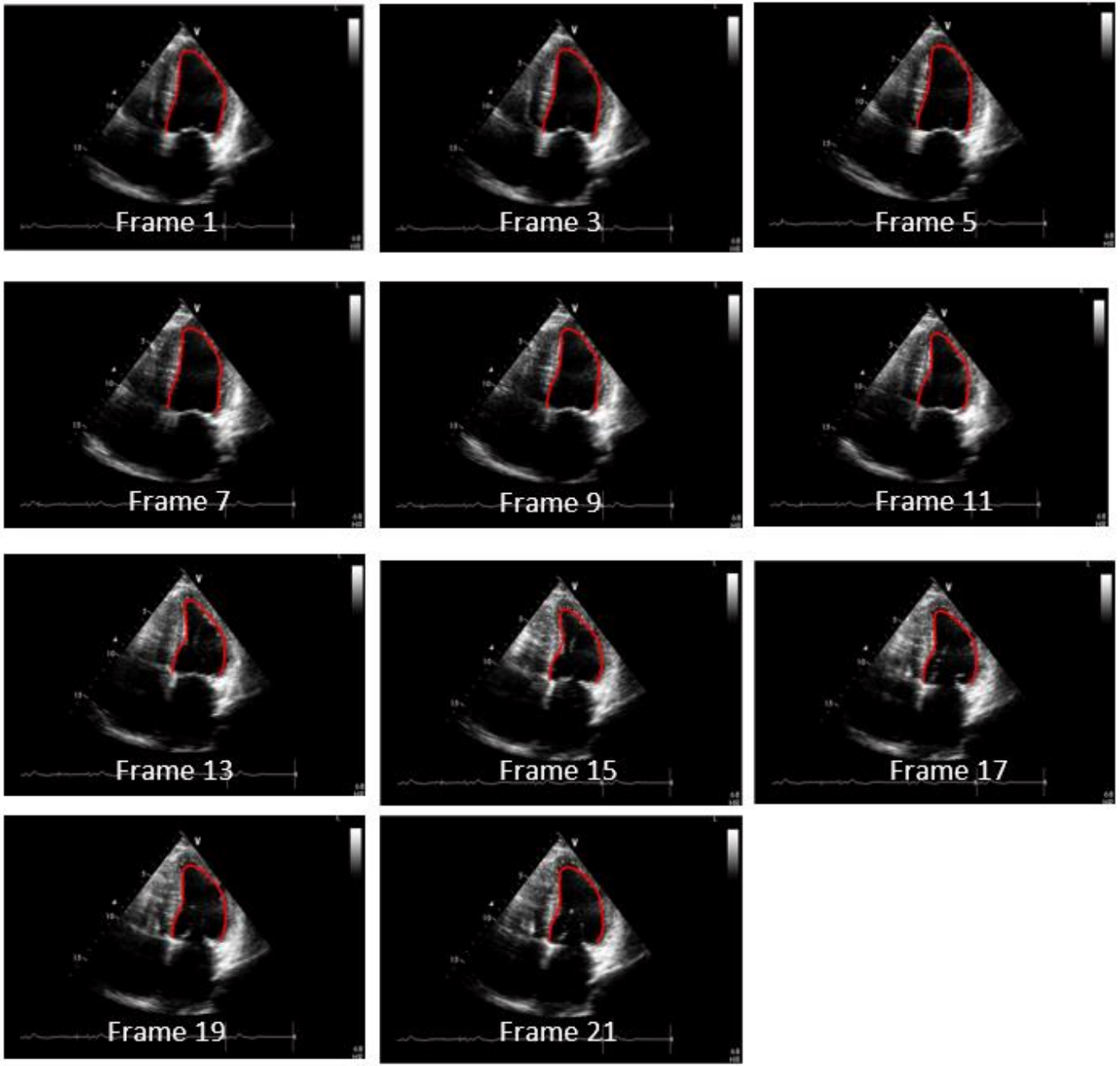

**Figure 16:** Initial guess results for every other frame in one clip

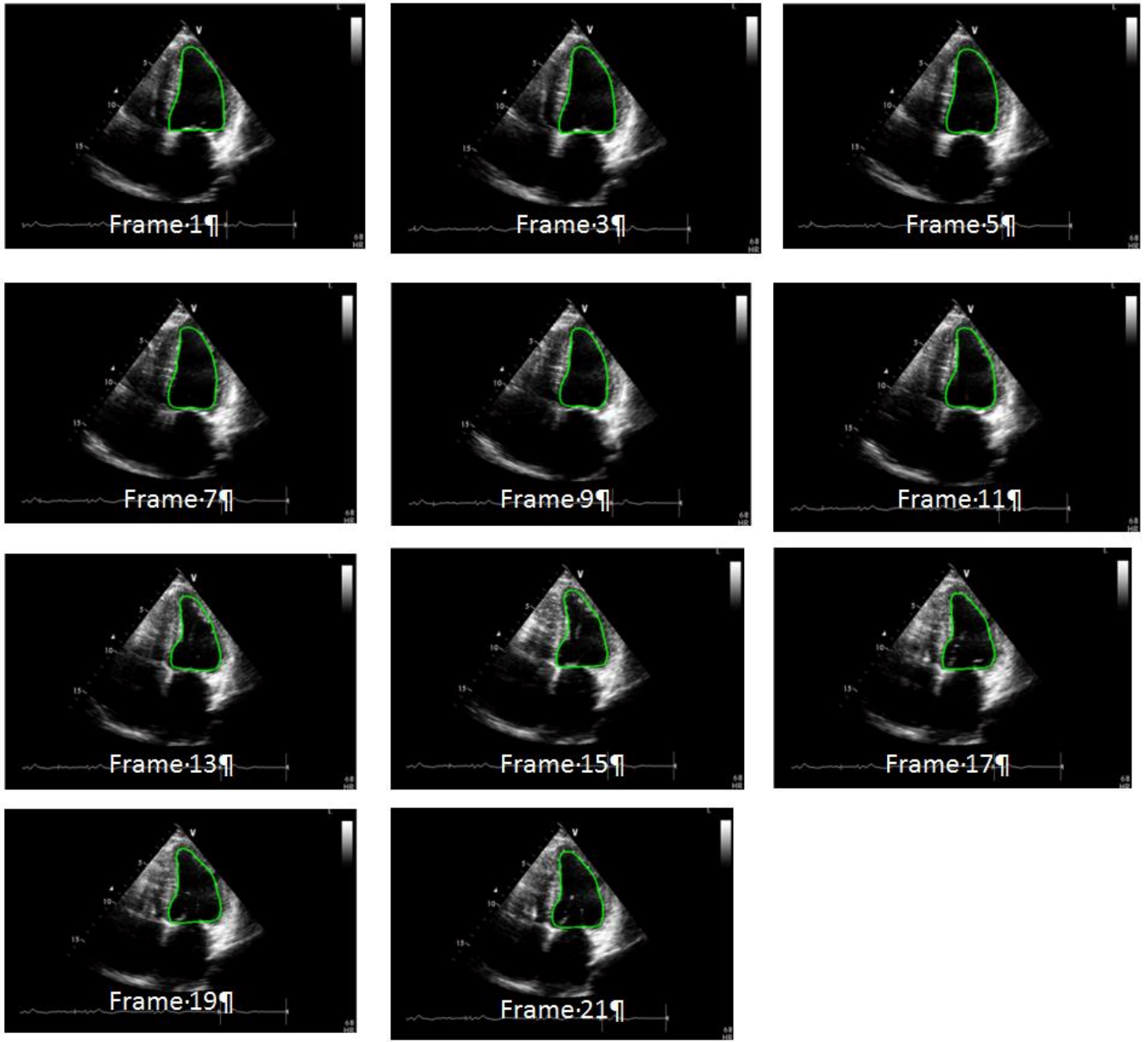

**Figure 17:** Final segmentation results for every other frame in one clip

**Study limitations**

Our method has shown good results but has a few limitations. Low image quality is naturally an obstacle in achieving accurate automatic segmentation, but in some acceptable quality ultrasound clips manual intervention of the user may still be required for determining the correct valve corner positions and\or the apex. Also, tracking the valve corners through a single cardiac cycle has proven to be robust, but in tracking through longer clips that may no longer be the case. The small and negligible tracking error accumulated between each frame may add up to a significant error in longer clips and therefore require modifications in the tracking algorithm or limit the segmentation to single heart beats.

V. CONCLUSION

To conclude, an algorithm for the segmentation of the left ventricle is proposed, in which the main anatomical features of the left ventricle are detected first and then used for defining an initial guess of the shape of the ventricle. An active contour algorithm can then be used for fine tuning of the contour. The results showed that the left ventricular contour found using our method successfully followed the endocardial border.

ACKNOWLEDGMENT

We would like to thank the Dept. of Cardiology, Pneumology and Angiology at the University Hospital RWTH Aachen, Germany, and the Dept. of Cardiology, at the Oslo University Hospital, Rikshospitalet, Oslo, Norway for providing us with the data.

This study was supported by the Magnetton Program, the Chief Scientist, Ministry of Industry and Commerce of Israel. We are grateful for this support.

REFERENCES

1. Friedland N, Adam D. Automatic ventricular cavity boundary detection from sequential ultrasound images using simulated annealing. IEEE transactions on medical imaging 1989, 8(4),344-353

2. Herlin IL, Giraudon G, Performing segmentation of ultrasound images using temporal information. Proceedings CVPR '93.

3. Herlin IL, Berziat D, Giraudon G, Nguyen C, Graffigne C, Segmentation of echocardiographic images with Markov random fields, Lecture notes in Computer Science 1994, 801/1994, 200-206

4. Mikic I, Krucinski S, Thomas JD. Segmentation and tracking in echocardiographic sequences: active contours guided by optical flow estimates. IEEE transactions on medical imaging 1998, 17(2) 274-284

5. Sun W, Cetin M, Chan R, Reddy V, Holmvang G, Chandar V, Willsky A. Segmenting and Tracking the left ventricle by learning the dynamics in cardiac images. Lecture notes in computer science 2005, 3536/2005,553-565

6. Coppini G, Poli R, Valli G, Recovery of the 3-D shape of the left ventricle from Echocardiographic Images, IEEE transactions on medical imaging (1995), 14(2) 301-317

7. Mignotte M, Meunier J, Tardif JC, Endoardial boundary estimation and tracking in echocardiographic Images using deformable templates and markov random fields. Pattern Analysis & Applications 2001, 4:256-271

8. Orderud F, A framework for real time left ventricular tracking in 3D+T echocardiography, using nonlinear deformable contours and kalman filter based tracking, computers in cardiology 2006;33:125-128

9. Orderud F, Hansgard J, Rabben SI, Real time tracking of the left ventricle in 3D echocardiology using a state estimation approach, Med Image Comput Comput Assist Interv. 2007;10(pt 1)858-65

10. Osher S, Sethian JA, Fronts propagating with curvature dependent speed: Algorithms based on Hamilton-Jacobi Formulations. Journal of computational physics 1988,79, 12-49

11. Lin N, Yu Weichuan, Duncan JS, Combinative Multi-scale level set framework for echocardiographic Image segmentation, Lecture notes in Computer science 2002, 2488/2002, 682-689

12. Corsi C, Saracino G, Sarti A, Lamberti C, Left ventricular volume estimation for real-time three dimensional echocardiography. IEEE transactions on medical imaging 2002, vol. 21 (9) ;1202-1208

13. Paragios N. A level set approach for shape-driven segmentation and tracking of the left ventricle. IEEE Trans Med Imaging. 2003;22:773-776

14. Angelini ED, Homma S, Pearson G, Holmes JW, Laine AF, Segmentation of real-time three-dimensional ultrasound for quantification of ventricular function: a clinical study on right and left ventricles. Ultrasound in Med & Biol. 2005,31(9) 1143-1158

15. Lankton S, Tannenbaum A, Localized Region based Active contours. IEEE transactions on Image processing 2008, 17(11), 2029-2039

16. Amini AA, Weymouth TE, JainRC. Using dynamic programming for solving variational problems in vision. IEEE Trans PAMI 1990;12:855-867

17. Sonka M, Hlavac V, Boyle R. Image processing, analysis and machine vision. 2nd ed. Pacific Grove, CA: International Thomson-Brooks/Cole, 1999;148-158

18. Üzümcü M, Van der Geest RJ, Swingen C, Reiber JHC, Lelievelds BPF, Time continuous tracking and segmentation of cardiovascular magnetic resonance images using multidimensional dynamic programming. Invest Radiol 2006;41:52-62

19. Nevo ST, Van Stralen M, Vossepoel AM, Reiber JHC, De Jong N, Van Der Steen AFW, Bosch JG, Automated Tracking of The Mitral Valve Annulus Motion in Apical Echocardiographic Images using Multidimensional Dynamic Programming, Ultrasound in Med & Biol, 2007;33(9):1389-1399

20. Rivaz H, Boctor E, Foroughi P, Zellars R, Fichtinger G, Hager G., Ultrasound elastography: a dynamic programming approach IEEE Trans Med Imaging. 2008 Oct;27(10):1373-7

21. Petrank Y, Huang L, O'Donnell M, Reduced peak-hopping artifacts in ultrasonic strain estimation using the viterbi algorithm. IEEE Trans Ultrason Ferroelectr Freq Control. 2009 Jul;56(7):1359-67

22. Wang Q, Song E, Jin R, Han P, Wang X, Zhou Y, Zeng J, Segmentation of lung nodules in computed tomography images using dynamic programming and multidirection fusion techniques. Acad Radiol. 2009 Jun;16(6):678-88.

23. Song E, Jiang L, Jin R, Zhang L, Yuan Y, Li Q., Breast mass segmentation in mammography using plane fitting and dynamic programming., Acad Radiol. 2009 Jul;16(7):826-35.

24. McCullough DP, Gudla PR, Harris BS, Collins JA, Meaburn KJ, Nakaya MA, Yamaguchi TP, Misteli T, Lockett SJ, Segmentation of whole cells and cell nuclei from 3-D optical microscope images using dynamic programming. IEEE Trans Med Imaging. 2008 May;27(5):723-34

25. Abd-Elmoniem KZ Abou-Bakr A, Kadah YM. Real-time speckle reduction and coherence enhancement in ultrasound imaging via nonlinear anistropic diffusion. IEEE Trans Biomed Engineering 2002 Sep;49(9):997-1014.

APPENDIX

Using DP for finding the endocardial border in each of the left and right parts of the left ventricular image, is done the following way: first, two matrices in the size of the image (left or right part) are defined. $W$ – a matrix for keeping the accumulated weights, and $F$ – a matrix for keeping the location from which the curve passing through a pixel located at ($x,y$), has arrived. The weight matrix is given the initial value of 0 in all locations, except in the apex location, which is given a very high bonus (large negative value) to make sure that the path passes through it.

The weight matrix values in the first row are updated as follows:

(1) $W(x,1) = W(x,1) + fg * G(x,1), \forall x \in (x, y = 1)$

where G is the gradient image and *fg* is a negative factor that multiplies the gradient value, so that the higher the gradient is, the smaller it's contribution will be to the accumulating weight, as we are looking for the path with minimal accumulated weight.

The weight matrix values and $F$ matrix values in the second row are then updated in steps described in equations 2-4:

(2) $W_{temp}(x,2) = W(x',1) + fdx * (|x - x'|), \forall x' \in (x', y = 1) \ \& \ |x - x'| < DX$

where $W_{temp}$ is a temporary weight vector which holds all the possible weights that paths arriving from a location x' in the first row, could contribute to a location $x$ in the second row, *fdx* is a positive weight factor for un continuity of the path, which is described by |x-x'|, and *DX* is the maximal number of pixels in the x direction we would expect the path to skip between rows

(3) $W(x,2) = W(x,2) + fg * G(x,2) + \min(W_{temp}), \forall x \in (x, y = 2)$

(4) $F(x,2) = x_{min}, \forall x \in (x, y = 2, W_{temp}(xmin) = \min(W_{temp}))$

where $x_{min}$ is the $x$ location of the minimal contributor to $W(x,2)$, or in other words, the location from which the path at ($x,y$) arrived.

Starting from the third row and on, updating the weight matrix and the F matrix is done as follows:

(5) $W_{temp}(x,y) = W(x', y-1) + fdx * (|x - x'|) + fd2x * |x + F(x', y-1) - 2 * x'|$
$\forall x' \in |x - x'| < DX$

where *fd2x* is a positive weight factor for un-smoothness of the path and the expression following it is the backwards second derivative of the path that expresses its smoothness. The *x* location of the path at *y-2* is given in *F(x',y-1)*. The weight and *F* matrices are updated as follows:

(6) $W(x,y) = W(x,y) + fg * G(x,y) + \min(W_{temp}), \forall x \in (x, y \geq 3)$

(7) $F(x,y) = x_{min}, \forall x \in (x, y \geq 3, W_{temp}(x_{min}) = \min(W_{temp}))$

In order to find the likeliest path P that starts at the apex and ends at the valve corner, we set the last value of the path as the x position of the valve:

(8) $P(n) = x_{valve}$

The matrix F at point x_valve points to the x position of the path at row n-1:

(9) $P(n-1) = F(x_{valve})$

Then:

(10) $P(n-2) = F(P(n-1))$

And so on.